\documentclass[runningheads]{llncs}

\usepackage{eccv}
\usepackage{eccvabbrv}

\usepackage{graphicx}
\usepackage{booktabs}
\usepackage[accsupp]{axessibility}  
\usepackage{hyperref}
\usepackage{orcidlink}
\usepackage{multirow}

\begin{document}

\title{Open-set object detection: towards unified problem formulation and benchmarking} 

\titlerunning{Open-set OD: towards unified problem formulation and benchmarking}

\author{Hejer Ammar \and
Nikita Kiselov \and
Guillaume Lapouge \and
Romaric Audigier}

\authorrunning{H. Ammar et al.}

\institute{Université Paris-Saclay, CEA, List, F-91120, Palaiseau, France
\email{\{firstname.lastname\}@cea.fr}}

\maketitle

\begin{abstract}

In real-world applications where confidence is key, like autonomous driving, the accurate detection and appropriate handling of classes differing from those used during training are crucial. Despite the proposal of various unknown object detection approaches, we have observed widespread inconsistencies among them regarding the datasets, metrics, and scenarios used, alongside a notable absence of a clear definition for unknown objects, which hampers meaningful evaluation. To counter these issues, we introduce two benchmarks: a unified \textit{VOC-COCO} evaluation, and the new \textit{OpenImagesRoad} benchmark which provides clear hierarchical object definition besides new evaluation metrics. Complementing the benchmark, we exploit recent self-supervised Vision Transformers \cite{dinov2} performance, to improve pseudo-labeling-based OpenSet Object Detection (OSOD), through OW-DETR$^{++}$. State-of-the-art methods are extensively evaluated on the proposed benchmarks. This study provides a clear problem definition, ensures consistent evaluations, and draws new conclusions about effectiveness of OSOD strategies.

  
\end{abstract}

\section{Introduction} \label{sec:intro}

Object Detection (OD) is a fundamental task in several applications of computer vision. Usually, corresponding models are trained to detect a closed set of object classes. However, in real life, they may encounter objects from new classes 
\textit{not seen} or \textit{not labelled} 
in the training dataset. We call these objects \textit{unknown} or \textit{Out Of Distribution} (OOD). In some applications, where safety and confidence are key, not confusing these objects with the known classes or even being able to detect and recognise them as unknowns is crucial. For example, an undetected or a misclassified unknown object can be dangerous for self-driving cars. 

While OOD-OD approaches \cite{VOS, SIREN} aim to correctly classify already detected objects as a known class or unknown, two other related topics focus also on the ability of the model to localise the unknowns. First, \textit{OpenSet Object Detection} (OSOD) approaches \cite{OpenDet, robotics} seek to actively localise all unknown objects without falsely categorising them as known classes. Second, \textit{Open-World Object Detection} (OWOD) methods \cite{OWOD, OW-DETR, MAVL, randbox} try to handle a continuously expanding \textit{world} of object classes. At different time steps, the model receives new sets of known classes and should be able to detect old, new and unknown classes changing through time. The present work focuses on OSOD.

While studying different unknown OD approaches, several problems have been raised. First, we noticed that researchers employ distinct sets of training and evaluation data, and inconsistent evaluation metrics. Moreover, different scenarios are used without a clear distinction between them. In fact, unknown objects can be present in the training dataset but not labeled (\textit{unlabeled}), or completely absent (\textit{unseen}). This makes it impossible to consistently compare all approaches. Another issue is the ambiguous definition of the \textit{unknown} object: what exactly do we want to detect? In this work, we address the initial problem by establishing a unified benchmark using the widely-adopted VOC \cite{VOC} and COCO \cite{COCO} datasets, along with the previously defined metrics. We demonstrate that this evaluation is incomplete due to the absence of unknown objects definition. To tackle this challenge, we adapt the known super-class concept introduced in \cite{OSOD-III} to multiple super-classes in the same dataset for a more general and realistic application. To this end, we introduce \textit{OpenImagesRoad}, a novel benchmark 
made of road images from OpenImages with refined annotations \cite{OpenImages}. Leveraging the extensive dataset hierarchy, we provide a clear definition of unknown objects. Besides, we propose suitable evaluation metrics. This enables a comprehensive evaluation of methods, assessing both their accuracy and proficiency in detecting known and unknown objects. These well-defined benchmarks, scenarios, baselines and problem characterisations should provide clearer and fair settings for analysing outcomes of future OSOD works.

Finally, we propose to exploit the capabilities of recent self-supervised Vision Transformer DINOv2 \cite{dinov2} to improve the state-of-the-art (SOTA) pseudo-labeling based unknown OD approach OW-DETR \cite{OW-DETR}. The proposed methods, OW-DETR$^{+}$ and OW-DETR$^{++}$ bring great improvements 
by leveraging the self-supervised pre-training of ViTs. OW-DETR$^{++}$ allows important interpretations about the capabilities of each family of approaches according to the used scenario. 

Our contributions can be summarized as follows:
\begin{itemize}
    \item We introduce a new benchmark: OpenImagesRoad. It provides a clear and comprehensive definition of the \textit{unknown} in the context of OD. This definition, alongside new evaluation metrics, enable to accurately assess OSOD method performances \footnote{All splits, annotations and evaluation codes are available at \url{https://kalisteo.cea.fr/index.php/free-resources/}.}.
    \item We introduce a new pseudo-labeling method: OW-DETR$^{++}$. It leverages self-supervised pre-trained vision transformers and achieves best results among pseudo-labeling methods.
    \item We overcome the literature inconsistencies by comparing them on the same benchmarks, namely the unified VOC-COCO and OpenImagesRoad benchmarks. Thus, we bring fresh perspective on the abilities and limits of different approaches depending on the used scenario. 
\end{itemize}

\section{Related Work}

\subsection{Background} \label{background}

Acquiring a confident object detector that is aware of the unknown is tackled by multiple methods under different settings and different names.  

\noindent\textbf{OOD} \textbar{} \textbf{O}ut-\textbf{O}f-\textbf{D}istribution detection was firstly associated with image classification. Latest approaches include post-hoc methods such as KNN with threshold-based criterion \cite{KNN} or rectifying output activations at an upper limit \cite{ReAct}. Other methods are ad-hoc and require specific re-training of the model. It includes modifying the loss function \cite{logitnorm}, ensembling and scoring rules \cite{DeepEnsemble} or even using external unlabelled datasets \cite{UDG}, GAN-generated images \cite{GanSyntehsis, OpenGan} or mixup augmentations \cite{pixmix} to represent the unknown. Naturally, these methods can be applied to OD, during the classification process. Detected objects are simply reclassified as known or unknown. Some approaches propose combining both custom loss and data generation in the feature space \cite{VOS}, while others \cite{SIREN} add a distance-based loss in the feature domain. However, those methods do not seek to improve the model's ability to localise unknown objects.

\noindent\textbf{OSOD} \textbar{} \textbf{O}pen-\textbf{S}et \textbf{O}bject \textbf{D}etection approaches aim to both localise and correctly classify the knowns and the unknowns. In this setting, the questions "what do we consider as unknown objects?" is fundamental. In the literature, this problem is often ill-posed as some works define knowns and unknowns as instances from the same dataset \cite{generalized-survey}, whereas others \cite{unified-survey} argue that this is the matter of class definition and not differences in the datasets. On the other hand, \cite{towardsOSR} groups unknowns into sub-groups of related unknown objects. Finally,  \cite{OSOD-III} defines the unknowns as all classes under a known super-class. However, it only considers one super-class per dataset, which is usually not the case in real-world applications. Whatever the problem definition, OSOD approaches \cite{OpenDet, robotics} propose custom loss functions and contrastive clustering for better known-unknown separability. Specifically, OpenDet \cite{OpenDet} identifies unknowns by differentiating between high and low-density regions in the latent space. It promotes compact features for known classes, thereby expanding low-density regions for unknowns. It also optimises the unknown probability based on the uncertainty of predictions. 

\noindent\textbf{OWOD} \textbar{} \textbf{O}pen-\textbf{W}orld \textbf{O}bject \textbf{D}etection is a setting introduced in ORE \cite{OWOD}, where the goal is to deal with expanding sets of classes through time. Instead of just localising and identifying known and unknown objects, OWOD systems aim also to continuously learn and incorporate new classes as they encounter them, without forgetting the old ones. First, ORE \cite{OWOD} adapts Faster-RCNN \cite{FasterRCNN} to the OWOD setting by integrating contrastive clustering and employing an automatic pseudo-labeling mechanism, leveraging RPN proposals with the highest objectness scores for potential unknown objects. Yet, this approach requires some labeled unknown samples for proper calibration, which violates the OOD hypothesis. OW-DETR \cite{OW-DETR} develops the idea of pseudo-labeling by using the most activated feature map regions as unknown objects. MAVL \cite{MAVL} exploits the capacities of multi-model ViTs trained with aligned image-text pairs to create a class agnostic object detector which pseudo-labels the unknowns resulting from the \textit{"all objects"} text query. Finally, RandBox \cite{randbox} learns the unknowns using random proposals as pseudo-labels to cover new distributions.

\subsection{SOTA inconsistencies and problem ambiguities}

\textbf{Methods inconsistencies:} Studying the aforementioned methods, we found a lot of nuances which can effect the models performances and usability in real-world conditions. Tab. \ref{SOTA-Comparaison} summarises a structured comparison of the most recent approaches, encapsulating their key characteristics, the strategies they adopt and the differences among them. First, the settings and goals of OOD-OD, OSOD and OWOD approaches are different such as explained in sec. \ref{background}. Moreover, some methods use an external supervision on other datasets, which can violate the open-set setting making the comparison unfair with other fully unsupervised methods on the unknowns. In addition, ORE \cite{OWOD} requires labeled unknown data for calibration (\textit{Unk. label}), while several methods need a calibration on a validation set to function optimally (\textit{Pre-calibration}). Furthermore, to detect the unknown, some methods are based on pseudo-labeling, where others rely on re-estimating detections using a learned distribution. Finally, note that there is no comparison along all previous SOTA approaches using the same benchmarks.

\begin{table*}[t!]{}\centering \small
\caption{Key characteristics of SOTA unknown detection methods.}
  \begin{center}
{\scalebox{0.8}{
\begin{tabular}{l|c|c|c|c|c|c}
\textbf{Method}  & \textbf{Year} &  \textbf{Problem type} & \parbox{2cm}{\centering \baselineskip=10pt \textbf{External \\ supervision}}  & \textbf{Unk. label} & \textbf{Pre-calibration} & \parbox{2cm}{\centering \baselineskip=10pt \textbf{Unk. det. \\ technique}}  \\
\hline
SIREN \cite{SIREN} & 2022  & OOD & $\checkmark$ & $\times$ & $\checkmark$ & Probability-based \\
VOS \cite{VOS} & 2022 & OOD & $\times$ & $\times$ & $\checkmark$ & Probability-based \\
OpenDet \cite{OpenDet} & 2022 & OSOD & $\times$ & $\times$ & $\times$ &  Probability-based \\
ORE \cite{OWOD}  & 2021 & OWOD & $\times$ & $\checkmark$ & $\checkmark$ &  Probability-based  \\
MAVL \cite{MAVL} & 2022  & OWOD & $\checkmark$ & $\times$ & $\times$ & Pseudo-labeling \\
OW-DETR \cite{OW-DETR} & 2022  & OWOD & $\times$ & $\times$ & $\times$ & Pseudo-labeling \\
RandBox \cite{randbox} & 2023 & OWOD & $\times$ & $\times$ & $\times$ & Pseudo-labeling \\
\end{tabular}}}
 \label{SOTA-Comparaison}
  \end{center}
\end{table*}

\noindent\textbf{The definition of training scenarios: }The use of different benchmarks results in the non-separation between the \textit{unlabeled} vs. the \textit{unseen} scenarios. In fact, while for some settings unknown objects can be present in the training dataset but not labeled, they are completely absent for others. \cite{dhamija2020overlooked} noticed that the VOC-COCO benchmark is part of the \textit{unlabeled} scenario which they called \textit{mixed unknowns} but they do not propose nor benchmark on a totally \textit{unseen} scenario. 

\noindent\textbf{The definition of unknown objects:} Another big ambiguity lies in the definition of unknown objects. In fact, for supervised OD, the goal is to detect a well defined set of classes both present and labeled in the training dataset. However, the term \textit{unknown} can encompass all entities that fall outside the realm of the recognized or the \textit{known}. In this case, the distinction between foreground and background (\eg are trees and buildings considered as unknown objects or background) and between objects and their parts (\eg do we want to detect a person as one object or to detect the different body parts and clothes as different objects) is unclear. The task of deciding what to detect and what to ignore without any prior annotation becomes complex, making a consistent evaluation impossible. Hosoya et al. \cite{OSOD-III} introduce a new definition of \textit{objects} named OSOD-III, by breaking down the challenge hierarchically. It proposes a class hierarchy wherein unknown objects are defined as all objects belonging to the same super-class as the known objects. For example, if the known classes are animal species, we expect the model to detect all other animal species as unknowns, arguing that they share visual similarities. While \cite{OSOD-III} provides a great first effort at rectifying OSOD ill-posedness, only single super-class benchmarks are proposed. Distinction between unlabeled and unseen is also not explicit. 

\noindent\textbf{Proposed metrics: } Previous OOD-OD, OSOD, and OWOD studies have employed diverse metrics for evaluation. First, False Positive Rate at true positive rate equals 95\% (\textbf{FPR95}) was inherited from OOD image classification and was largely adopted in OOD-OD. It evaluates the capacity of the model to correctly classify the detected objects, but it ignores its ability to localise the unknowns. Later, Absolute Open-Set Error (\textbf{A-OSE}) and Wilderness Impact (\textbf{WI}) have been proposed to evaluate OSOD performance \cite{DropoutSample, dhamija2020overlooked}. A-OSE quantifies the absolute number of unknown objects mistakenly classified as known. WI, on the other hand, measures the proportion of A-OSE among all known detections. Note that WI does not make sense in a pure OOD domain, where no known objects are present. Moreover, these two metrics are insufficient for comprehensive OSOD evaluation, indeed, they focus on a single type of error \ie detecting known objects as unknown, essentially ignoring the others. Furthermore, they operate at specific confidence thresholds and thus overlook the precision-recall trade-off, a fundamental aspect of OD \cite{OSOD-III}. Consequently, these metrics might provide an incomplete or skewed perspective on the model's performance. Moreover, to evaluate the localisation capacity, \textbf{U-Recall} was defined as the ratio of unknown detected objects by the total number of unknown objects. Although U-Recall penalises absent unknown detections, it does not consider the precision and the confidence scores. Finally, as OSOD-III \cite{OSOD-III} delivers a clean object definition, it proposes to use the Unknown Average Precision (\textbf{$AP_{unk}$}), along with the usual mean Average Precision on the known objects ($mAP_{k}$). By encompassing the recall-precision curve, $AP_{unk}$ gives a more relevant evaluation of unknown detection. However, it evaluates jointly the localisation and the classification performances \eg if a model detects all unknown objects but classifies them as known classes under the same super-class, the $AP_{unk}=0$.  

\section{New Problem Definition and Unified Benchmarks}

In this section, we propose new unified benchmarks, provide scenarios separation and adapt the \cite{OSOD-III} problem definition to richer contexts to consistently evaluate SOTA OSOD and OWOD approaches in the same OSOD framework, as OWOD methods can be also considered as OSOD approaches with a supplementary option \ie dealing with an expanding world of new classes through time. 

Similarly to usual supervised OD methods, we define the OSOD training dataset as $D_{train}=\{(x_i, y_i), x_i\in X, y_i \in Y\}$, where $x_i$ is an input image and its corresponding labels $y_i=\{(c_j, {b}_j)\}_{j=1}^{N_i}$. $N_i$ denotes the number of labeled objects in this image, $b$ are the bounding box coordinates for each object and $c$ are their corresponding class labels. We note $\mathcal{K} = \{1, \dots, C\}, c \in \mathcal{K}$ as the set of classes introduced during training \ie known classes. The goal during testing on $D_{test}$, is to jointly detect objects from known classes $\mathcal{K} $ and identify objects from other unknown classes $\mathcal{U} $, ensuring that they are not misclassified into $\mathcal{K} $. For practical purposes, as it is challenging to enumerate all possible unknown classes, we consider them as one class $\mathcal{U} = \{C+1\} $. 

We first propose a unified benchmark and test splits on a combination of the well-known Pascal-VOC \cite{VOC} and MS-COCO \cite{COCO} datasets in sec. \ref{voccoco}. In sec. \ref{scenarios}, we separate between two scenarios which have been overlooked by previous works. Then, we propose a new richer benchmark derived from the OpenImages datasets \cite{OpenImages} for a better \textit{unknown} definition (see sec. \ref{openimages}). Finally, we suggest suitable evaluation metrics for a complete and consistent evaluation of different approaches in sec. \ref{metrics}.

\subsection{New unified OSOD benchmark} \label{voccoco}

While a significant number of OWOD and OSOD approaches use a combination of Pascal-VOC \cite{VOC} and MS-COCO \cite{COCO}, they each introduce unique splits and protocols, preventing methods comparison. Recognizing this issue, we first propose a consolidated split that introduces minimal deviations from the original VOC and COCO configurations. To our knowledge, we are the first to benchmark latest OWOD and OSOD methods on a singular platform. We advocate for our approach as a foundational step towards unifying methodologies and ensuring consistent evaluation. To evaluate the known and unknown detection capacities separately and jointly, our evaluation is segmented into three distinct splits:
\begin{itemize}
    \item $D_{test,ID} = \{x_i, \forall c_j \in y_i, c_j  \in \mathcal{K}\}$  is the set of test images containing only known objects (pure ID);
    \item $D_{test,OOD} = \{x_i, \forall c_j \in y_i, c_j  \in \mathcal{U}\}$  is the set of test images containing only unknown objects (pure OOD);
    \item $D_{test,all} = D_{test}$ is the set of all test images, containing known, unknown, or both kinds of objects.
\end{itemize}
For training, we used the usual VOC trainval split such as \cite{dhamija2020overlooked}, which has $20$ classes in $14K$ images. For testing, we used both VOC and COCO test sets while adopting the aforementioned splits for finer analysis. In fact, COCO dataset encompasses the same $20$ classes from VOC along with additional $60$ non-VOC classes. These are aggregated under a single \textit{unknown} label. For this benchmark, $D_{test,ID}$ is the standard VOC test split, $D_{test,all}$ is the combination of VOC and COCO test splits, and $D_{test,OOD}$ contains only COCO images where no object of the $20$ known classes is present. 

\begin{table}[t]
    \centering
    \caption{Confusion matrix of content and labels of train and test datasets and scenarios definition.} 
    \includegraphics[width=0.65\textwidth]{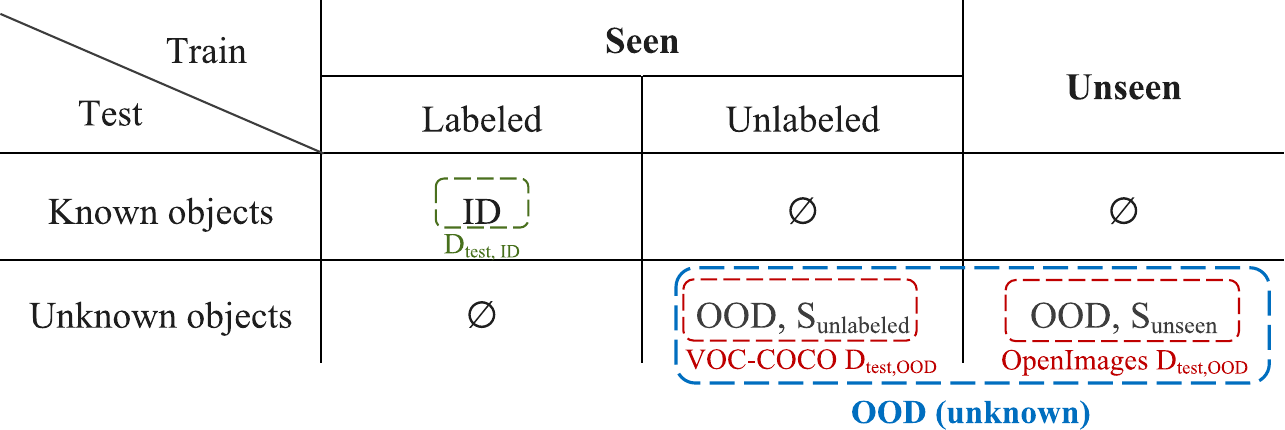}
    \label{tab:scenarios}
\end{table}

\subsection{Training scenarios} \label{scenarios}

While studying unknown detection settings, two different scenarios can be defined, such as presented in tab. \ref{tab:scenarios}:
\begin{itemize}
    \item The \textit{unlabeled} scenario ($S_{unlabeled}$): where unknown objects can be seen but unlabeled in the $D_{train}$. In this scenario, the unknown objects are not fully OOD. Note that the VOC-COCO benchmarks are actually part of $S_{unlabeled}$ as a considerable number of VOC images contain objects from the $60$ unknown COCO classes while being not labeled (see appendix for illustrations).
    \item The \textit{unseen} scenario ($S_{unseen}$): where unknown objects have not been seen during training. In this case, unknown objects are completely new and OOD. Note that, to the best of our knowledge, no such benchmark has been proposed yet, despite being the more realistic one, where the goal is to correctly handle completely new objects \eg an electronic scooter can be an unknown object unseen in training autonomous vehicle scenes.
\end{itemize}

\subsection{OpenImagesRoad: A richer unseen hierarchical benchmark} \label{openimages}

\noindent\textbf{Hierarchical object definition:} We think OSOD-III \cite{OSOD-III} is a step in the right direction as it defines an object as all classes belonging to a same single known super-class. Similarly, we employ a hierarchical approach to define what an object is, but extend the idea to multiple super-classes that can be present in a single dataset. 
We believe that such hypothesis is practical, realistic and holistic for contemporary applications such as autonomous driving where a wide range of super-classes are encountered and where unknown objects should be handled appropriately. In this case, we consider that the dataset is hierarchically annotated; each class can be categorised under one super-class. We define $SC=\{SC_l, 0 \le l \le L\}$ as the set of super-classes in the dataset. We divide each super-class $SC_l$ into a subset of known and unknown classes. In this case, the goal is to accurately detect all known instances, and all novel objects within predetermined super-classes. Every other detection that falls outside the defined $SC$ will be considered as a false detection. This rich and clear definition of what we want to detect allows a consistent evaluation of the unknown detection abilities.  In practice, we have chosen the OpenImages dataset as a base for our own evaluation protocol thanks to its annotation and class hierarchy definition. 

\noindent\textbf{OpenImages \cite{OpenImages}:} is a large dataset with $1.9M$ images. The classes are structured in a hierarchical manner, resembling a tree, with secondary classes branching out of primary super-classes. However, such as demonstrated by BigDetection \cite{cai2022bigdetection}, the initial OpenImages dataset presents several problems such as overlapping annotations or redundant class representations, \eg an object can be annotated twice such as \textit{skyscraper} and \textit{tower}. This can create ambiguity in object definitions, making it hard to differentiate between known and unknown entities. To address this challenge, we harness the benefits of BigDetection \cite{cai2022bigdetection} that refines the OpenImages annotations, removes redundant or overlapped bounding boxes and maps the OpenImages categories to a unified label space, which is based on the commonly accepted WordNet \cite{wordnet} hierarchy, which covers a wide range of environments. 

\noindent\textbf{OpenImagesRoad:} To create a smaller and more specific benchmark close to real-life applications, we introduce the \textit{OpenImagesRoad} dataset which is a subset of OpenImages containing only road images and using the BigDetection annotations and hierarchy. To this end, we select every image that contains at least one \textit{vehicle} or \textit{street sign} object and do not contain any indoor object (under super-classes \textit{home appliance}, \textit{plumbing fixture}, \textit{office supplies}, \textit{kitchenware}, \textit{furniture}, \textit{bathroom accessory}, \textit{drink}, \textit{food}, \textit{cosmetics}, \textit{personal care}, \textit{medical equipment}, \textit{musical instrument} and \textit{computer electronics}). \textit{OpenImagesRoad} contains a total of $228153$ images and $1120348$ objects. The hierarchical depth of the BigDetection annotations is fully employed. In fact, we split each super-class into two distinct sets: the most frequent $50\%$ classes having at least $60$ instances are considered as known classes, and the rest (less frequent $50\%$ classes or having less than $60$ examples) are unknown. This resulted into $50$ known class, and $113$ unknown class all grouped under the label \textit{unknown}. Note that we deleted all classes corresponding to object parts such as \textit{clothing}. We can imagine another homogeneous dataset derived from OpenImages \eg indoor images or in the wild images, thanks to the exhaustive and hierarchical annotation of OpenImages allowing the known object super-class concept. 

Such as described in sec. \ref{scenarios}, the VOC-COCO benchmark is under the $S_{unlabeled}$ scenario and no previous benchmark was proposed under the more challenging $S_{unseen}$ scenario, despite being more realistic. To this end, we propose to split the OpenImagesRoad benchmark according to $S_{unseen}$ where images containing any unknown objects are only used for test or validation (more details are provided in the appendix). In fact, contrary to VOC dataset, and thanks to the extensive annotation of the OpenImages dataset, we can assure that no unknown object is present in our selected training split. 

\subsection{Metrics} \label{metrics}

While $AP_{unk}$ offers a great insight into the model performance to accurately detect the unknown in a well defined benchmark such as OpenImagesRoad, it can hide some model characteristics by combining the localisation and classification tasks. For example, if a model localises well all unknown objects but fails to classify them as unknown $AP_{unk}=0$. To provide a finer analysis, we propose two new metrics. 

First, we adopt a class agnostic average precision as a localization-focused metric which we call $AP_{all}$. It evaluates the localisation performances of the model for both known and unknown, while considering the precision-recall trade-off, regardless of the classification results. $AP_{all}$ is equal to the average precision, computed when considering all predictions and all ground-truth annotations as elements of the same class. 

Second, we define a new low granularity metric: $AP_{sc}$. It evaluates the classification and localization performance following the super-class hierarchy. In fact, some models can confuse unknown objects with a close known class belonging to the same super-class (e.g. a \textit{helicopter} can be wrongly classified as an \textit{airplane} instead of an unknown). We consider such behaviour acceptable in some use-cases and define $AP_{sc}$ as the average precision when considering the right super-class category. Note that $AP_{sc}$ can be only computed on $D_{test,OOD}$ where all objects are unknown. 
Let us consider $GT_u$ as a ground-truth unknown object having a class $c_u$ under a super-class $SC_l$. We define $TP_{sc}$, $FN_{sc}$ and $FP_{sc}$ as true positives, false negatives and false positives for super-class respectively, to compute the recall-precision curve of $AP_{sc}$ as a usual average precision using the objectness scores.  A detection is considered a true positive ($TP_{sc}$ is incremented) if its Intersection over Union ($IoU$) with $GT_u$ is greater than $0.5$ and is classified as unknown $\mathcal{U}$ or as a known class $c_k$ under the same super-class ($ c_k \in SC_l$). It is a false positive ($FP_{sc}$ is incremented), if it does not overlap with any remaining $GT_u$ ($IoU<0.5$) under the same-super class. Finally, a $GT_u$ is considered not detected ($FN_{sc}$ is incremented) if no detection overlaps with it or if it is detected as a known class $c_k$ under a different super-class ($c_k \notin SC_l$). 

\section{Can We Improve Pseudo-labeling OSOD?}

Pseudo-labeling approaches are an efficient way to approach OSOD and OWOD. We find them particularly useful for the simple knowledge distillation they allow from self-supervised models.
However, many pseudo-labeling techniques such as ORE \cite{OWOD} or MAVL \cite{MAVL}, are based on supervised learning paradigms, requiring additional external annotated data which is out of the scope of the work presented here. Moreover, these methods are constrained by the narrow definition of what constitutes an object, derived primarily from the labeled data they have been trained on. On the other hand, in recent years, self-supervised Vision Transformers (ViT) demonstrated tremendous capacities in representation learning without any prior annotated knowledge. 

In this section, we present our efforts to push the capacities of the SOTA OW-DETR \cite{OW-DETR} method by leveraging DINOv2 \cite{dinov2} self-supervised features for better pseudo-labeling. Because we consider that our contribution is a refinement of OW-DETR, we call the resulting methods OW-DETR$^{+}$ and OW-DETR$^{++}$.

\subsection{Background: OW-DETR \cite{OW-DETR}}

\textbf{OW-DETR} | \textbf{O}pen-\textbf{W}orld \textbf{DE}tection \textbf{TR}ansformer \cite{OW-DETR} is an adaptation of the object detector D-DETR (Deformable DETR) \cite{d_detr} tailored for OWOD. At the core of the D-DETR framework, the multi-scale features of an image are extracted from the backbone: a ResNet \cite{Resnet} that was self-supervised on ImageNet \cite{imagenet}. Subsequently, these features 
are directed to a transformer-based encoder-decoder mechanism which outputs detections. In OW-DETR, the decoder introduces pseudo-labeling for unknown objects by evaluating unmatched proposals against ground-truth annotations. The $K_u$ proposals having the most activated activation map regions, are labeled as \textit{unknown}. The model learns to detect both known and unknown objects jointly. Note that $K_u$ is a fixed hyper-parameter for all images in the training dataset.

\subsection{Pseudo-labeling with Self-supervised ViT} \label{pseudo-label}

\begin{figure*}[t]
    \centering
    \includegraphics[width=1\textwidth]{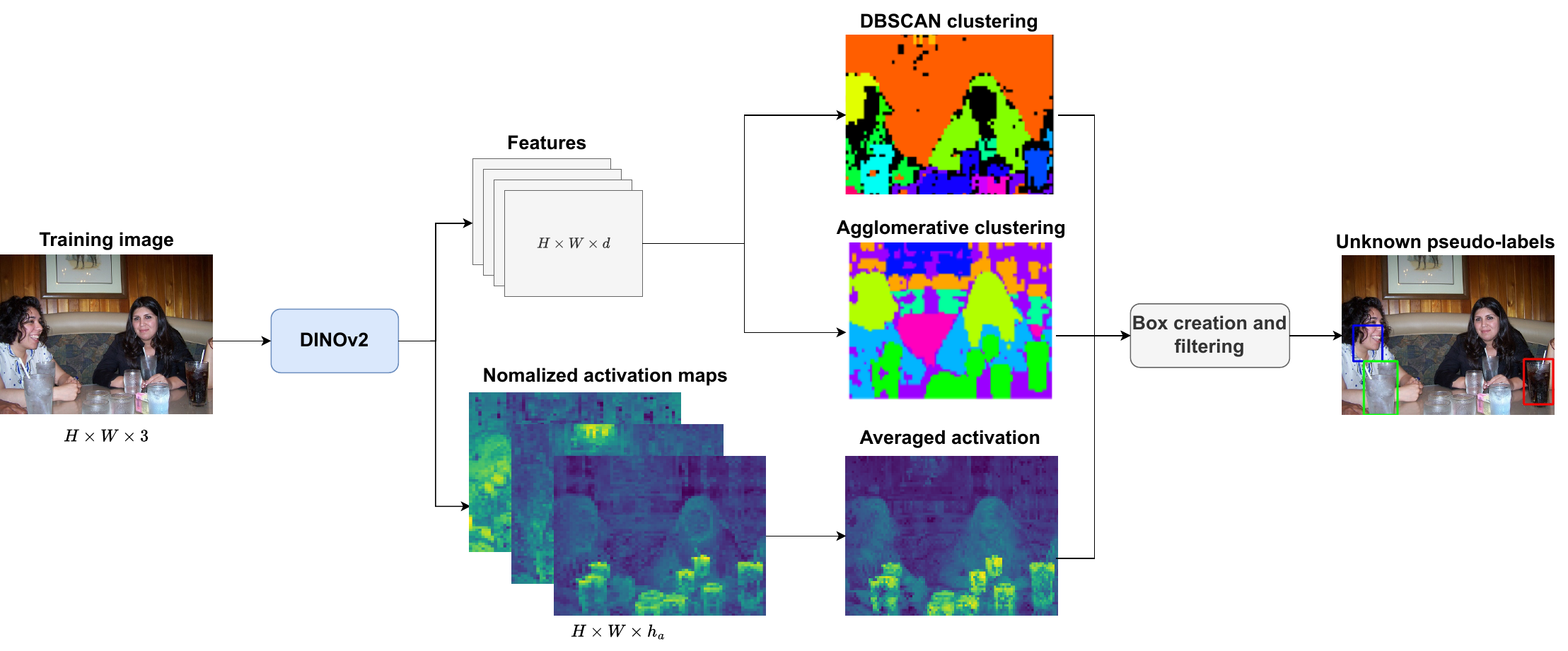}
    \caption{OW-DETR$^{++}$ pseudo-labeling pipeline.
    }
    \label{fig:overall_arch}
\end{figure*}

\textbf{OW-DETR$^{+}$} | In order to improve OW-DETR, by leveraging pre-trained ViT features, a first step is to replace Resnet activation maps used for pseudo-labeling by DINOv2 \cite{dinov2} activation maps. We call this variant OW-DETR$^{+}$. Note that DINOv2 is only used for the activation map extraction for the unknown pseudo-labeling while ResNet continues to be the backbone of the detection method. 

\noindent\textbf{OW-DETR$^{++}$} | To further improve {OW-DETR$^{+}$}, we propose the novel pseudo-labeling pipeline illustrated in fig. \ref{fig:overall_arch}. We first propose to filter out the background. Given an input image passed through the DINOv2 backbone, we obtain a feature map $F \in \mathbb{R}^{H \times W \times d}$, where each pixel is represented by a $d$-dimensional feature vector. Subsequently, we apply DBSCAN \cite{dbscan} with predefined $eps$ and $min\_samples$ on these features since it can automatically assign the number of clusters. This algorithm provides a great ability to separate between the foreground and the background regions. Hence, we filter out the biggest cluster, likely to be the background, especially for OD benchmarks where usually single objects are not omnipresent in the image. Second, we aim to find different objects in the image by localising the regions having similar and homogeneous semantics. However, DBSCAN offers very coarse semantic clusters. Thus, we apply agglomerative clustering on the same features, refine the clusters boundaries and minimize noisy representations using morphological operations such as erosion and dilation. The feature space is then segmented into $AC$ distinct clusters, according to a distance threshold $d_{th}$. This clustering helps in determining coherent object-like structures. Note that each cluster represents potentially a class, as different objects from the same class have usually similar features.
We aim to keep only regions having high objectness by using the DINOv2 activation maps. In fact, $h_a$ attention maps $A_i =\{a_{h,w}, 1 \leq h \leq H,  1 \leq w \leq W \}$ are derived from the ViT backbone, where $h_a$ represents the number of attention heads. Each pixel value $a_{h,w}$ for each attention map is normalized to prevent outlier activations: $a_{h,w} = \frac{a_{h,w}}{\sum_{l=1}^{H} \sum_{k=1}^{W} a_{l,k}}$. These maps are averaged to obtain: $A_{avg} = \frac{1}{h_a} \sum_{i=1}^{h_a} A_i$. Using $A_{avg}$, we compute the average attention activation $A_{ac}$ for each cluster $ac$. Intuitively, higher activations correspond to regions containing objects. Given this insight, we only keep clusters having an average activation greater than the mean average attention of the whole image ($A_{ac}>mean(A_{avg})$). We then search for spatially isolated instances in these clusters and surround the largest $N$ regions with bounding boxes. Finally, we apply Non-Maximum Suppression (NMS) to the generated bounding boxes and filter out boxes that overlap with ground truth boxes ($IoU>T$). Note that in our case, we can have zero unknown object if top activated regions overlap with ground truth objects. We can also have a large number of unknown objects, if different clusters are largely activated. Following the aforementioned method, pseudo-labels for unknown objects are pre-computed in an offline manner for each image of the training dataset. It is important to note that this is a lengthy process. The D-DETR architecture is then trained end-to-end to detect objects from $|\mathcal{K}|+1$ classes by predicting $|\mathcal{K}|+1$ softmax probabilities  ($|\mathcal{K}|$ known class and one unknown class).

\section{Experiments}

\subsection{Datasets and evaluation metrics}

We deliver results on the two proposed benchmarks VOC-COCO and OpenImagesRoad, for the three test splits such as described in sections \ref{voccoco} and \ref{openimages}.

For known classes, we report the standard mean average precision ($mAP_k$) on both benchmarks. 
For unknown classes, we only report A-OSE and U-Recall on the VOC-COCO benchmark, since unknown objects are not clearly defined. We then report U-Recall, $AP_u$, $AP_{all}$ and $AP_{sc}$ (when applicable) on the OpenImagesRoad benchmark, thanks to the unknown object definition.

\subsection{Implementation Details} 

To deliver a consistent and fair evaluation, we use the official implementations of all tested methods. To maintain fairness, for ORE \cite{OWOD}, we calibrated the distribution exclusively on $D_{train}$ with only labeled known objects. For MAVL \cite{MAVL}, we use the public pre-trained model for pseudo-labeling our dataset, and then use these pseudo-labels to train ORE.

For, OW-DETR$^{+}$ and OW-DETR$^{++}$ we use the same implementation details as OW-DETR, but change the pseudo-labeling process as described in sec.  \ref{pseudo-label}. For both variants, the pseudo-labels are extracted using ViT-S/14 DINOv2 pre-trained model instead of the Resnet-50 \cite{Resnet} pre-trained on ImageNet \cite{imagenet} used for OW-DETR. Note that this configuration is chosen to keep comparable the number of parameters ($23M$ for Resnet-50 and $22M$ for ViT-S/14). 384-dimentional feature vectors are extracted from each image ($d=384$), and $h_a=14$ attention heads are used. Finally, for OW-DETR$^{++}$, we choose the DBSCAN parameters permitting the best foreground/background separation: $eps=7$ and $min\_samples=35$. For agglomerative clustering, $d_{th}$ is set to $245$ and was defined as the minimum distance between clusters of known classes on VOC training dataset. In addition, we use the default parameters of agglomerative clustering set by the \textit{sklearn} library. We set $N=3$ as the maximum number of objects kept from the same unknown cluster and $T=0.3$ to filter pseudo-labels that overlap with ground truth boxes. Different ablation studies are in the appendix.

\subsection{VOC-COCO benchmark}

\begin{table}[t]
    \caption{VOC-COCO benchmarking results. Best results are in \textbf{bold}, second best are \underline{underlined}.}
    \centering
\small
{\scalebox{0.8}{
\begin{tabular}{l|c|ccc|cc}
    \midrule
    Method & \multicolumn{1}{c|}{$D_{test,ID}$} & \multicolumn{3}{c|}{$D_{test,all}$} & \multicolumn{2}{c}{$D_{test,OOD}$} \\ 
    \cmidrule(lr){2-2} \cmidrule(lr){3-5} \cmidrule(lr){6-7}
    & $mAP_{k \uparrow}$ & $mAP_{k \uparrow}$ & U-Recall$_{\uparrow}$ & A-OSE$_{\downarrow}$  & U-Recall$_{\uparrow}$ & A-OSE$_{\downarrow}$ \\ \midrule
    OpenDet \cite{OpenDet} & \bf{82.7} & \bf{63.4} & 14.5 & \textbf{3712} & 33.4 & \textbf{1536} \\ 
    ORE \cite{OWOD}& 80.0  & 60.8 & 7.5 & 8279 & 21.5 & 3541 \\ 
    ORE (MAVL \cite{MAVL}) & \underline{80.2} & 60.6 & \textbf{50.2} & \underline{6008} & \bf{68.1} & \underline{2884} \\
    RandBox \cite{randbox} & 66.2 & 48.9 & 20.5 & 44224 &  44.1 & 15972  \\
    OW-DETR \cite{OW-DETR} & 78.8 & 60.5 & 13.4 & 41643 & 28.5 & 21022 \\ 
    OW-DETR$^{+}$ & 75.4 & 57.8 & 28.4 & 28354 & 53.2 & 13458 \\
    \midrule
    OW-DETR$^{++}$ & 80.0 & \underline{61.2} & \underline{36.0} & 13470 & \underline{56.1} & 5685 \\ 
    \midrule
\end{tabular}}}
    \label{tab:voc_coco}
\end{table}

We provide a performance evaluation of previous SOTA OSOD and OWOD methods, alongside the different contributions that we brought to OW-DETR on the proposed VOC-COCO benchmark in tab. \ref{tab:voc_coco}.

For the known objects, OpenDet outperforms all methods with the highest $mAP_k$. Note that, for all benchmarked methods, $mAP_k$ is higher for $D_{test,ID}$ than $D_{test,all}$, where the presence of both known and unknown objects may lead to more confusion.

However, the VOC-COCO benchmark makes it difficult to extract interesting conclusions regarding the unknown detection.
Indeed, a high U-Recall only indicates that unknown predictions have been successfully placed over unknown objects. However, it does not indicate proper detection performance, which should be measured as a recall-precision trade-off for different confidence thresholds. A-OSE is a first step to measure this trade-off, but only covers classification performance. Since both metrics are computed at fixed confidence thresholds, which can have different meanings for different detectors, it is impossible to properly assess the unknown detection performance.
That being said, ORE (MAVL) then OpenDet and finally OW-DETR$^{++}$ seem to be the best performing methods when analysing the U-Recall \vs A-OSE trade-off. However, it is worth mentioning that MAVL is a pseudo-labeling method that uses an external annotated dataset of text-image pairs to train the class-agnostic detector. Thus, in this $S_{unlabeled}$ scenario, OW-DETR$^{++}$ outperforms previous pseudo-labeling methods not using external supervision. Finally, we can observe the significant unknown detection improvements brought to OW-DETR through OW-DETR$^{+}$ and then OW-DETR$^{++}$. Indeed, leveraging a richer ViT representation to achieve better pseudo-labeling results in paramount improvements.

\subsection{OpenImagesRoad benchmark} 

\begin{table*}[t!]{}\centering \small
\caption{OpenImagesRoad benchmarking results. Best results are in \textbf{bold}.}
{\scalebox{0.8}{
\begin{tabular}{l|c|cccc|cccc}
    \midrule
    Method & \multicolumn{1}{c|}{$D_{test,ID}$} & \multicolumn{4}{c|}{$D_{test,all}$} & \multicolumn{4}{c}{$D_{test,OOD}$} \\ 
    \cmidrule(lr){2-2} \cmidrule(lr){3-6} \cmidrule(lr){7-10}
    & $mAP_{k \uparrow}$ & $mAP_{k \uparrow}$ & $AP_{u \uparrow}$ & $AP_{all \uparrow}$ & U-Recall$_{\uparrow}$ & $AP_{u \uparrow}$ & $AP_{all \uparrow}$ & $AP_{sc \uparrow}$ & U-Recall$_{\uparrow}$  \\ \midrule
    OpenDet \cite{OpenDet} & 20.3 & 18.4 & \textbf{9.3} & 48.6 & 71.5 & \textbf{45.4} & 32.5 & 11.7 & 82.3 \\ 
    RandBox \cite{randbox} & 16.3 & 14.8 & 3.1 & 3.9 & 49.0 & 26.0 & 31.4 & 4.1 & 60.0   \\
    OW-DETR \cite{OW-DETR} & 18.4 & 16.7 & 0.2 & 23.9 & 31.6 & 1.6 & 18.8 & 11.5 & 39.1 \\ 
    OW-DETR$^{+}$ & 19.0 & 17.4 & 0.7 & 28.9 & 69.3 & 4.5 & 36.8 & 15.7 & 81.6 \\
    \midrule
    OW-DETR$^{++}$ & \textbf{21.9} & \textbf{19.9} & 3.1 & \textbf{49.4} & \textbf{74.5} & 9.0 & \textbf{56.4} & \textbf{37.9} & \textbf{84.7} \\ 
    \midrule
\end{tabular}}}
\label{tab:openimages}
\end{table*}

We compare in tab. \ref{tab:openimages} performances of OpenDet \cite{OpenDet}, RandBox \cite{randbox} and OW-DETR \cite{OW-DETR} alongside its different improvements. ORE is unreported as it shows poor performance in unknown detection in tab. \ref{tab:voc_coco}. For fairness, ORE (MAVL \cite{MAVL}) is also unreported as it uses external supervision to extract pseudo-labels. 

\begin{figure*}[t]
    \centering
    \includegraphics[width=1\textwidth]{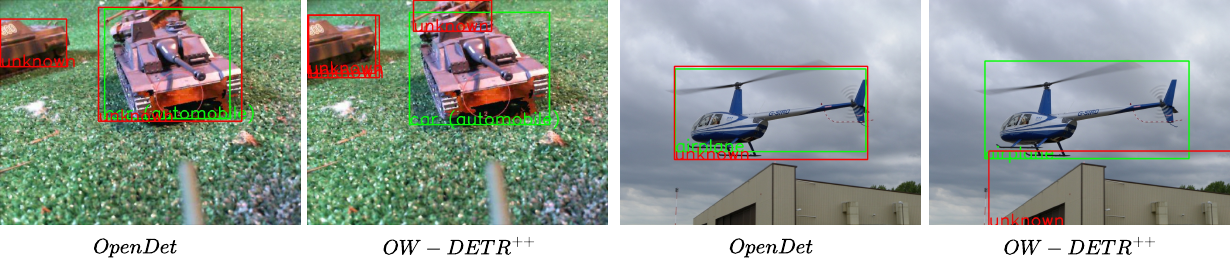}
    \caption{Samples of OpenDet \cite{OpenDet} and OW-DETR$^{++}$ detections. Only unknown objects are present in the images. In green (resp. red), box predicted as known (resp. unknown).}
    \label{fig:samples}
\end{figure*}

Interesting performance trends emerge depending on the nature of each method. Indeed, pseudo-labeling methods such as RandBox, OW-DETR,  OW-DETR, OW-DETR$^{+}$ and OW-DETR$^{++}$ all have lower $AP_{u}$ than the contrastive method OpenDet which indicates that contrastive methods seem to learn a broader unknown representation which is good for classification. On the other hand, OW-DETR$^{++}$  have higher $AP_{all}$ and $AP_{sc}$ than OpenDet, demonstrating that this pseudo-labeling based method seems more apt for object localization. The higher $AP_{sc}$ also indicates a good ability of pseudo-labeling methods to understand high granularity classification but at the same time, an inability to understand smaller grained differences between known and unknown as shown by the lower $AP_{u}$. The surprising drop between $AP_{u}$ and $AP_{all}$ for OpenDet can be explained by frequent double detections of unknown objects: as unknown and known objects. Fig. \ref{fig:samples} illustrates the behaviours of OpenDet and OW-DETR$^{++}$, further samples of all methods are included in the appendix.
 
Our proposed pseudo-labeling improvements on OW-DETR, namely OW-DETR$^{+}$ and OW-DETR$^{++}$ bring impressive performance gains. Note that no knowledge on unknown objects is transferred from DINOv2 since it is only used to pseudo-label the training images and that no unknown object is present in these images for the \textit{unseen} scenario. Indeed, OW-DETR$^{++}$ achieves best performance for known OD ($mAP_{k}$), object localization $(AP_{all}$), and high granularity classification unknown OD ($AP_{sc}$). However, despite the drastic performance increase, OpenDet remains better for correctly detecting unknown objects and classifying them as unknown ($AP_{u}$).

To conclude, the proposed OpenImagesRoad benchmark  permitted to shine new light on the detection performance of these methods. It appears that there is not a single best method among those we tested. Rather, the choice of a method can be driven by the envisaged use-case. If unknown objects should be both correctly detected and classified as unknown, the contrastive learning based OpenDet detector appears to be an appropriate choice, especially for such $S_{unseen}$ scenario. Conversely, if the correct localisation of unknown objects is paramount, and coarse classification is sufficient, then OW-DETR$^{++}$ seems best indicated. It is also worth noting that pseudo-labeling techniques are expected to perform better in the $S_{unlabeled}$ scenario where unknown objects are potentially seen and pseudo-labeled during training. Indeed, we remark that in tab. \ref{tab:voc_coco} (VOC-COCO/$S_{unlabeled}$), OW-DETR$^{++}$ seems to be relatively more preferment compared to Opendet, than in tab. \ref{tab:openimages} (OpenImagesRoad/$S_{unseen}$).

\section{Conclusions}

In this work, we propose a new unified OSOD problem formulation and benchmarking. We introduced the OpenImagesRoad hierarchical dataset which permits OSOD performance evaluation in realistic, well defined conditions. We also proposed the unified VOC-COCO benchmark for fast, although limited, benchmarking of OSOD and OWOD methods.
We brought improvements to the pseudo-labeling baseline OW-DETR, namely OW-DETR$^{+}$ and OW-DETR$^{++}$, and compared these methods alongside main SOTA approaches.
Both new benchmarks and methods evaluation allowed us to shine new light on OSOD performance. First, pseudo-labeling and contrastive methods each have advantages and drawbacks, therefore the choice of the method depends on the chosen use-case. Second, methods are sensitive to the learning scenario with pseudo-labeling methods thriving in $S_{unlabeled}$ scenarios when compared to $S_{unseen}$.

\section*{Acknowledgements}
Funded by the European Union. Views and opinions expressed are however those of the author(s) only and do not necessarily reflect those of the European Union or the European Commission. Neither the European Union nor the granting authority can be held responsible for them.

This work benefited from the FactoryIA supercomputer financially supported by the Ile-deFrance Regional Council.

\clearpage  

%
%
\bibliographystyle{splncs04}
\bibliography{egbib}

\begin{thebibliography}{10}
\providecommand{\url}[1]{\texttt{#1}}
\providecommand{\urlprefix}{URL }
\providecommand{\doi}[1]{https://doi.org/#1}

\bibitem{cai2022bigdetection}
Cai, L., Zhang, Z., Zhu, Y., Zhang, L., Li, M., Xue, X.: Bigdetection: A large-scale benchmark for improved object detector pre-training. In: Proceedings of the IEEE/CVF Conference on Computer Vision and Pattern Recognition. pp. 4777--4787 (2022)

\bibitem{imagenet}
Deng, J., Dong, W., Socher, R., Li, L.J., Li, K., Fei-Fei, L.: Imagenet: A large-scale hierarchical image database. In: 2009 IEEE Conference on Computer Vision and Pattern Recognition. pp. 248--255 (2009). \doi{10.1109/CVPR.2009.5206848}

\bibitem{dhamija2020overlooked}
Dhamija, A., Gunther, M., Ventura, J., Boult, T.: The overlooked elephant of object detection: Open set. In: WACV (2020)

\bibitem{SIREN}
Du, X., Gozum, G., Ming, Y., Li, Y.: Siren: Shaping representations for detecting out-of-distribution objects. Advances in Neural Information Processing Systems  \textbf{35},  20434--20449 (2022)

\bibitem{VOS}
Du, X., Wang, Z., Cai, M., Li, Y.: Vos: Learning what you don't know by virtual outlier synthesis. arXiv preprint arXiv:2202.01197  (2022)

\bibitem{VOC}
Everingham, M., Van~Gool, L., Williams, C.K., Winn, J., Zisserman, A.: The pascal visual object classes (voc) challenge. IJCV  \textbf{88}(2),  303--338 (2010)

\bibitem{wordnet}
Fellbaum, C. (ed.): WordNet: An Electronic Lexical Database. Language, Speech, and Communication, MIT Press, Cambridge, MA (1998)

\bibitem{OW-DETR}
Gupta, A., Narayan, S., Joseph, K., Khan, S., Khan, F.S., Shah, M.: Ow-detr: Open-world detection transformer. In: Proceedings of the IEEE/CVF Conference on Computer Vision and Pattern Recognition. pp. 9235--9244 (2022)

\bibitem{dbscan}
Hahsler, M., Piekenbrock, M., Doran, D.: {dbscan}: Fast density-based clustering with {R}. Journal of Statistical Software  \textbf{91}(1),  1--30 (2019). \doi{10.18637/jss.v091.i01}

\bibitem{OpenDet}
Han, J., Ren, Y., Ding, J., Pan, X., Yan, K., Xia, G.S.: Expanding low-density latent regions for open-set object detection. In: Proceedings of the IEEE/CVF Conference on Computer Vision and Pattern Recognition. pp. 9591--9600 (2022)

\bibitem{Resnet}
He, K., Zhang, X., Ren, S., Sun, J.: Deep residual learning for image recognition. In: CVPR. pp. 770--778 (2016)

\bibitem{pixmix}
Hendrycks, D., Zou, A., Mazeika, M., Tang, L., Li, B., Song, D., Steinhardt, J.: Pixmix: Dreamlike pictures comprehensively improve safety measures. In: Proceedings of the IEEE/CVF Conference on Computer Vision and Pattern Recognition. pp. 16783--16792 (2022)

\bibitem{OSOD-III}
Hosoya, Y., Suganuma, M., Okatani, T.: Rectifying open-set object detection: A taxonomy, practical applications, and proper evaluation (2022), \url{https://api.semanticscholar.org/CorpusID:254070139}

\bibitem{OWOD}
Joseph, K., Khan, S., Khan, F.S., Balasubramanian, V.N.: Towards open world object detection. In: Proceedings of the IEEE/CVF conference on computer vision and pattern recognition. pp. 5830--5840 (2021)

\bibitem{OpenGan}
Kong, S., Ramanan, D.: Opengan: Open-set recognition via open data generation. In: Proceedings of the IEEE/CVF International Conference on Computer Vision. pp. 813--822 (2021)

\bibitem{OpenImages}
Kuznetsova, A., Rom, H., Alldrin, N., Uijlings, J., Krasin, I., Pont-Tuset, J., Kamali, S., Popov, S., Malloci, M., Kolesnikov, A., Duerig, T., Ferrari, V.: The open images dataset v4: Unified image classification, object detection, and visual relationship detection at scale. IJCV  (2020)

\bibitem{DeepEnsemble}
Lakshminarayanan, B., Pritzel, A., Blundell, C.: Simple and scalable predictive uncertainty estimation using deep ensembles. Advances in neural information processing systems  \textbf{30} (2017)

\bibitem{GanSyntehsis}
Lee, K., Lee, H., Lee, K., Shin, J.: Training confidence-calibrated classifiers for detecting out-of-distribution samples. arXiv preprint arXiv:1711.09325  (2017)

\bibitem{COCO}
Lin, T.Y., Maire, M., Belongie, S., Hays, J., Perona, P., Ramanan, D., Doll{\'a}r, P., Zitnick, C.L.: Microsoft coco: Common objects in context. In: ECCV. pp. 740--755. Springer (2014)

\bibitem{MAVL}
Maaz, M., Rasheed, H., Khan, S., Khan, F.S., Anwer, R.M., Yang, M.H.: Class-agnostic object detection with multi-modal transformer. In: European Conference on Computer Vision. pp. 512--531. Springer (2022)

\bibitem{DropoutSample}
Miller, D., Nicholson, L., Dayoub, F., S{\"u}nderhauf, N.: Dropout sampling for robust object detection in open-set conditions. In: ICRA (2018)

\bibitem{dinov2}
Oquab, M., Darcet, T., Moutakanni, T., Vo, H., Szafraniec, M., Khalidov, V., Fernandez, P., Haziza, D., Massa, F., El-Nouby, A., et~al.: Dinov2: Learning robust visual features without supervision. arXiv preprint arXiv:2304.07193  (2023)

\bibitem{FasterRCNN}
Ren, S., He, K., Girshick, R., Sun, J.: {Faster R-CNN}: Towards real-time object detection with region proposal networks. IEEE TPAMI pp. 1137--1149 (2017)

\bibitem{unified-survey}
Salehi, M., Mirzaei, H., Hendrycks, D., Li, Y., Rohban, M.H., Sabokrou, M.: A unified survey on anomaly, novelty, open-set, and out-of-distribution detection: Solutions and future challenges. arXiv preprint arXiv:2110.14051  (2021)

\bibitem{ReAct}
Sun, Y., Guo, C., Li, Y.: React: Out-of-distribution detection with rectified activations. Advances in Neural Information Processing Systems  \textbf{34},  144--157 (2021)

\bibitem{KNN}
Sun, Y., Ming, Y., Zhu, X., Li, Y.: Out-of-distribution detection with deep nearest neighbors. In: International Conference on Machine Learning. pp. 20827--20840. PMLR (2022)

\bibitem{randbox}
Wang, Y., Yue, Z., Hua, X.S., Zhang, H.: Random boxes are open-world object detectors. In: Proceedings of the IEEE/CVF International Conference on Computer Vision (ICCV). pp. 6233--6243 (October 2023)

\bibitem{logitnorm}
Wei, H., Xie, R., Cheng, H., Feng, L., An, B., Li, Y.: Mitigating neural network overconfidence with logit normalization. In: International Conference on Machine Learning. pp. 23631--23644. PMLR (2022)

\bibitem{UDG}
Yang, J., Wang, H., Feng, L., Yan, X., Zheng, H., Zhang, W., Liu, Z.: Semantically coherent out-of-distribution detection. In: Proceedings of the IEEE/CVF International Conference on Computer Vision. pp. 8301--8309 (2021)

\bibitem{generalized-survey}
Yang, J., Zhou, K., Li, Y., Liu, Z.: Generalized out-of-distribution detection: A survey. arXiv preprint arXiv:2110.11334  (2021)

\bibitem{towardsOSR}
Zheng, J., Li, W., Hong, J., Petersson, L., Barnes, N.: Towards open-set object detection and discovery. In: Proceedings of the IEEE/CVF Conference on Computer Vision and Pattern Recognition. pp. 3961--3970 (2022)

\bibitem{robotics}
Zhou, Z., Yang, Y., Wang, Y., Xiong, R.: Open-set object detection using classification-free object proposal and instance-level contrastive learning. IEEE Robotics and Automation Letters  \textbf{8}(3),  1691--1698 (2023)

\bibitem{d_detr}
Zhu, X., Su, W., Lu, L., Li, B., Wang, X., Dai, J.: Deformable detr: Deformable transformers for end-to-end object detection. In: ICLR (2021)

\end{thebibliography}
\end{document}